\documentclass[9pt,technote]{IEEEtran} 

\usepackage{amsmath,amsfonts,mathtools}
\usepackage{algorithm}
\usepackage{array}
\usepackage[caption=false,font=footnotesize]{subfig}
\usepackage{textcomp}
\usepackage{stfloats}
\usepackage{url}
\usepackage{verbatim}
\usepackage{graphicx}
\usepackage[nocompress]{cite}
\usepackage[T1]{fontenc}
\usepackage{algpseudocode}
\usepackage{upgreek}
\usepackage{bm}
\usepackage[cmintegrals]{newtxmath}

\newcommand{\btheta}{\ensuremath{\bm{\uptheta}}}
\newcommand{\bp}{\ensuremath{\mathbf{p}}}
\newcommand{\bH}{\ensuremath{\mathbf{H}}}

\begin{document}
\bstctlcite{IEEEexample:BSTcontrol}

\title{Self-Improving Interference Management Based on Deep Learning With Uncertainty Quantification}

\author{Hyun-Suk Lee, Do-Yup Kim, \IEEEmembership{Member, IEEE}, Kyungsik Min, \IEEEmembership{Member, IEEE}
\thanks{Hyun-Suk Lee is with the Department of Intelligent Mechatronics Engineering, Sejong University, Seoul 05006, South Korea (e-mail: hyunsuk@sejong.ac.kr).}
\thanks{Do-Yup Kim is with the Department of Information and Telecommunication Engineering, Incheon National University, Incheon 22012, South Korea (e-mail: doyup@inu.ac.kr).}
\thanks{Kyungsik Min is with the Department of Information and Telecommunications Engineering, The University of Suwon, Hwaseong 18323, South Korea (e-mail: kyungsik@suwon.ac.kr).}}

\markboth{Journal of \LaTeX\ Class Files,~Vol.~14, No.~8, August~2021}%
{Shell \MakeLowercase{\textit{et al.}}: A Sample Article Using IEEEtran.cls for IEEE Journals}


\maketitle

\begin{abstract}
This paper presents a groundbreaking self-improving interference management framework tailored for wireless communications, integrating deep learning with uncertainty quantification to enhance overall system performance.
Our approach addresses the computational challenges inherent in traditional optimization-based algorithms by harnessing deep learning models to predict optimal interference management solutions.
A significant breakthrough of our framework is its acknowledgment of the limitations inherent in data-driven models, particularly in scenarios not adequately represented by the training dataset.
To overcome these challenges, we propose a method for uncertainty quantification, accompanied by a qualifying criterion, to assess the trustworthiness of model predictions.
This framework strategically alternates between model-generated solutions and traditional algorithms, guided by a criterion that assesses the prediction credibility based on quantified uncertainties.
Experimental results validate the framework's efficacy, demonstrating its superiority over traditional deep learning models, notably in scenarios underrepresented in the training dataset.
This work marks a pioneering endeavor in harnessing self-improving deep learning for interference management, through the lens of uncertainty quantification.
\end{abstract}


\section{Introduction}

\IEEEPARstart{I}{n} the realm of wireless communications, extensive efforts have been dedicated to interference management, a critical challenge in enhancing system performance.
Specifically, many traditional algorithms have been devised to effectively address interference management optimization problems based on optimization theory, such as the weighted minimum mean squared error (WMMSE) and interference pricing algorithms \cite{papandriopoulos2009scale, shi2011iteratively, baligh2014cross}.
Nonetheless, the practical implementation of these algorithms still encounters many obstacles due to their substantial computational demands \cite{sun2018learning}.


To address this challenge, deep learning, like in other domains of wireless communications, has been extensively applied in interference management \cite{sun2018learning,shen2020graph}.
Rather than solving the problem directly, this approach involves training a deep learning model to predict the optimal solution to the problem specified by a given channel realization.
The model can be trained using a dataset consisting of channel realizations and their corresponding optimal solutions.
Once trained, the model can efficiently predict the optimal solution for any given channel realization, offering a cost-effective alternative to optimization-based algorithms.


However, this data-driven approach based on deep learning may not guarantee effective interference management in situations that deviate significantly from the given dataset.
Indeed, this limitation is inherent to data-driven approaches, since a single dataset cannot encompass all the diverse scenarios encountered in real-world systems \cite{abdar2021review}.
Hence, to ensure effective interference management, it is essential to assess the confidence level of the solution predicted by the deep learning model.
Then, based on this evaluation, optimization-based algorithms can be employed to complement the deep learning model while collecting additional data samples to improve model, enhancing its performance.

%


In this paper, we propose a self-improving interference management framework based on deep learning with uncertainty quantification.
Within this framework, we introduce a method to quantify the uncertainties associated with solutions made by the deep learning model.
Leveraging this quantified uncertainty, we devise a qualifying criterion to evaluate the trustworthiness of the model's solutions in terms of system performance.
Using this criterion, the framework selectively employs the model's solutions only when they are deemed trustworthy; otherwise, traditional algorithms are employed.
Through this iterative process, the framework collects data samples for situations that the model struggles with, subsequently allowing for self-improvement using the accumulated data.
We demonstrate via experimental results that our proposed framework is effective, not only for interference management but also for enhancing the deep learning model itself.
To the best of our knowledge, this paper represents the first endeavor to design a self-improving deep learning framework for interference management via uncertainty quantification in the field of deep learning.

\section{Deep Learning-Based Interference Management}

\subsection{System Model and Problem Formulation}

We consider a downlink network comprising $N$ pairs of single-antenna transceivers, each pair consisting of a base station (BS) and a corresponding scheduled user.
The set of BSs is defined as $\mathcal{N}=\{1,2,...,N\}$.
Let $h_{nn}$ denote the channel gain between BS $n$ and its scheduled user, and $h_{nm}$ the interference channel gain from BS $m$ to the scheduled user of BS $n$.
The channel gain matrix is defined as $\mathbf{H}\coloneqq[h_{ij}]_{i,j\in\mathcal{N}}$.
We assume that the channels main unchanged during each timeslot.
We denote the transmission power of BS $n$ as $p_n$ and define the vector of transmission powers as $\mathbf{p}\coloneqq[p_n]_{n\in\mathcal{N}}$.
Then, the signal to interference-plus-noise ratio (SINR) for the user of BS $n$ is given by
\begin{equation}
	\label{eqn:sinr}
	\textrm{SINR}_n \coloneqq \frac{\vert h_{nn} \rvert^2p_n}{\sigma_n^2+\sum_{m\in\mathcal{N},m\neq n}\lvert h_{nm}\rvert^2 p_m},
\end{equation}
where $\sigma_n^2$ is the noise power at that user.
The sum-rate of the network is given by
\begin{equation}
\label{eqn:weighted_sumrate}
R(\mathbf{H},\mathbf{p})=\sum_{n\in\mathcal{N}} \log(1+\textrm{SINR}_n).
\end{equation}
To maximize the weighted sum-rate, a power allocation problem for managing downlink interference is formulated as
\begin{equation}
\label{eqn:problem}
\max_{\mathbf{p}\in\mathcal{P}} ~ R(\mathbf{H},\mathbf{p}),
\end{equation}
where $\mathcal{P}\coloneqq[0,P^{\textrm{max}}]^N$ with $P^{\textrm{max}}$ being the maximum transmission power.
The optimal solution for a given $\mathbf{H}$ is denoted as $\mathbf{p}^*(\mathbf{H})$.

%
%

\subsection{Deep Learning for Interference Management}
\label{sec:dl_based_interference_management}

In the literature, a number of algorithms have been developed to address the interference management problem as formulated in \eqref{eqn:problem} with a given channel gain matrix $\mathbf{H}$ \cite{papandriopoulos2009scale,shi2011iteratively,baligh2014cross}.
Most of these algorithms employ an iterative approach, and their effectiveness has been demonstrated through simulation results and theoretical analyses.
However, implementing these algorithms in practical systems proves, in many cases, challenging due to their computational complexity.

\begin{figure}[!t]
\centering	
\subfloat[Training stage.]{\includegraphics[scale=0.6]{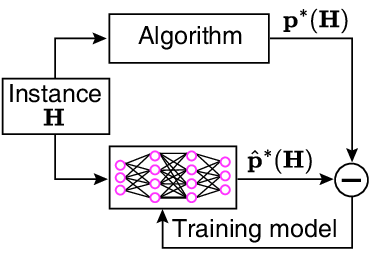}%
\label{fig:trad_training}}
\hfil
\subfloat[Testing stage.]{\includegraphics[scale=0.6]{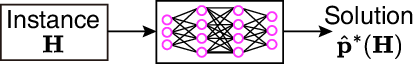}%
\label{fig:trad_test}}
\caption{A typical deep learning-based interference management approach.}
\label{fig:typical_deep_learning}
\end{figure}

To overcome this issue, recently, deep learning-based approaches to interference management have gained significant attention \cite{sun2018learning,shen2020graph}.
These approaches treat an interference management algorithm as a function $f(\mathbf{H})$ that takes $\mathbf{H}$ as input and yields $\mathbf{p}^*(\mathbf{H})$ as output.
Then, a deep neural network (DNN) model, $g_{\btheta}(\mathbf{H})$, is trained to approximate $f(\mathbf{H})$, where $\btheta$ represents the weights of the DNN model.
After training, the approximated solution $\hat{\mathbf{p}}^*(\mathbf{H})$ can be rapidly computed using the DNN model, requiring far fewer calculations than traditional algorithms.
Fig. \ref{fig:typical_deep_learning} illustrates a typical deep learning-based approach to interference management.
This approach involves generating a dataset from the target algorithm for given channel gain matrices, i.e., pairs of $(\mathbf{H},\mathbf{p}^*(\mathbf{H}))$.
Then, the DNN model is trained using this dataset, as shown in Fig. \ref{fig:trad_training}, and subsequently applied to interference management, as shown in Fig. \ref{fig:trad_test}.
This implies that the efficacy of the DNN model highly depends on the dataset used for training.
However, generating a comprehensive dataset that encompasses all possible channel conditions and corresponding power allocations is impractical.
Consequently, it is inherently challenging to ensure the DNN model's effectiveness in all conceivable interference scenarios, especially as some scenarios might exceed the generalization capabilities of deep learning and fall outside the scope of the training dataset.

\section{Self-Improving Interference Management Using Deep Learning}

\begin{figure}[!t]
\centering	
\subfloat[Training stage.]{\includegraphics[scale=0.6]{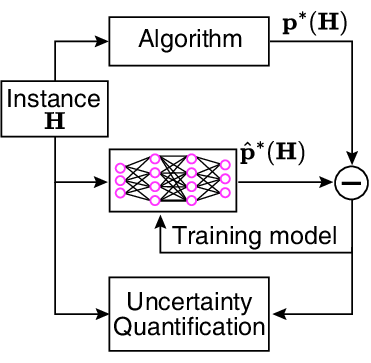}%
\label{fig:si_training}}
\hfil
\subfloat[Self-Improving stage.]{\includegraphics[scale=0.6]{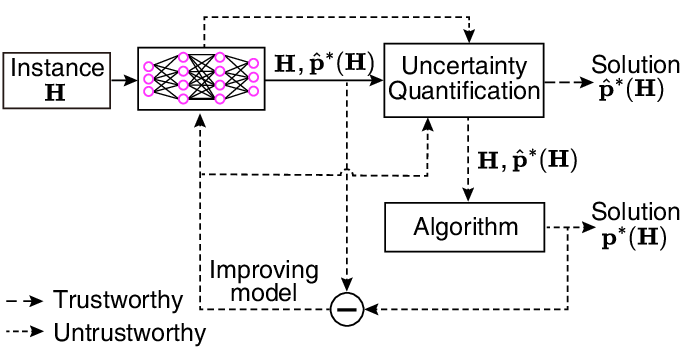}%
\label{fig:si_self_improving}}	
\caption{The proposed self-improving interference management framework using deep learning with uncertainty quantification.}
\label{fig:self_improving_deep_learning}
\end{figure}

In this section, we introduce the concept of uncertainty quantification in deep learning and its relevance to our context. We then present a proposed framework for self-improving interference management, which integrates deep learning with uncertainty quantification, as illustrated in Fig. \ref{fig:self_improving_deep_learning}.

\subsection{Concept of Uncertainty Quantification}

The quantification of predictive uncertainty in deep learning has been widely studied due to its role in assessing the trustworthiness of predictions.
%
%
Predictive uncertainty mainly stems from two sources: aleatoric and epistemic uncertainties \cite{abdar2021review}.
Aleatoric uncertainty is associated with inherent variabilities in the data distribution, rather than a characteristic of the DNN models.
As such, it is considered irreducible, meaning that it cannot be reduced through the collection of more data or improvements in the models.
In contrast, epistemic uncertainty stems from the limitations in DNN models, often due to insufficient training or inadequate experiences.
This type of uncertainty is typically prominent in regions of the input feature space that are not well-represented by the training dataset and is, therefore, reducible.
For more detailed theoretical background on uncertainty quantification, readers are referred to \cite{abdar2021review}.

Deep ensemble is one of the representative uncertainty quantification methods in deep learning \cite{lakshminarayanan2017simple}.
In uncertainty quantification, the output for a given input feature is treated as a random variable encompassing uncertainty.
Therefore, each DNN model in deep ensemble aims to approximate both the mean and variance of the output value of a given input feature, in contrast to typical DNN models that focus on approximating a single output value.
To effectively quantify uncertainties, deep ensemble involves training multiple models, each with distinct random initialization.
The prediction from these models are subsequently integrated to derive the final estimate.

\subsection{Uncertainty Quantification in Deep Learning-Based Interference Management}
\label{sec:UQ_ICIC}
We now describe an uncertainty quantification approach in deep learning-based interference management using deep ensemble.\footnote{In this paper, we use deep ensemble for quantifying uncertainty in deep learning due to its simplicity of implementation and adaptability to various types of NNs. However, other uncertainty quantification methods, such as Bayesian NN \cite{blundell2015weight} and conformal regression \cite{romano2019conformalized}, can also be applied within our proposed framework.}
To quantify uncertainty in interference management, we consider $M$ DNN models that jointly approximate the following distributional parameters of transmission power $p^*_n$ for a given channel gain matrix $\mathbf{H}$: the mean function $\mu_{n}(\mathbf{H})$ and the variance function $\sigma^2_{n}(\mathbf{H})$, where $\mu_{n}$ and $\sigma^2_{n}$ describe the solution $p^*_n(\mathbf{H})$ and the uncertainty of the estimate, respectively.
We denote the estimation of $\mu_n$ and $\sigma^2_n$ by the $m$th DNN model as $\hat\mu_{n,\btheta_m}$ and $\hat\sigma^2_{n,\btheta_m}$, respectively.
Note that we cannot simply use a mean squared error (MSE) loss function, a standard loss function for regression problems, as a loss function for training the models since the variance $\sigma^2_{n,\btheta_m}$ cannot be trained via the MSE.
Hence, to train the models for approximating both mean and variance, we employ a negative log-likelihood function as the loss function:
\begin{equation}
\label{eqn:loss_function}
\mathcal{L}_{n,\btheta}(\bH,p_n)=\frac{\log\hat\sigma^2_{n,\btheta}(\bH)}{2}+\frac{(p_n^*-\hat\mu_{n,\btheta}(\bH))^2}{2\hat\sigma^2_{n,\btheta}(\bH)}+c,
\end{equation}
where $c$ is a constant added to the loss function for practical reasons, such as numerical stability, without altering the fundamental behavior of the optimization process.
Then, by using the loss function, the DNN models with random initialization are trained by using given dataset.

We now estimate the uncertainty of the predicted solution $\hat\bp^*$ by averaging the predictions of $M$ DNN models, which implies that the individual distributions are ensembled as a uniformly-weighted mixture model.
For transmission power $p_n^*$, the averaged mean is given by $\hat{\mu}_n(\bH)=\frac{1}{M}\sum_{m=1}^M \hat{\mu}_{n,\btheta_m}(\bH)$, and the averaged variance is given by $\hat{\sigma}^2_n(\bH)=\frac{1}{M}\sum_{m=1}^M \left( \hat{\sigma}^2_{n,\btheta_m}(\bH)+\hat{\mu}^2_{n,\btheta_m}(\bH) \right)-\hat{\mu}_n$.
We can further decompose the variance as
\begin{equation}
\label{eqn:quantified_uncertainty}
\hat{\sigma}^2_n(\bH)=\underbrace{\frac{1}{M}\sum_{m=1}^M \hat{\sigma}^2_{n,\btheta_m}(\bH)}_{\textrm{Aleatoric variance}}+\underbrace{\frac{1}{M}\sum_{m=1}^M\hat{\mu}^2_{n,\btheta_m}(\bH) -\hat{\mu}_n}_{\textrm{Epistemic variance}},
\end{equation}
where the first term, the averaged estimated variance over all models, signifies aleatoric variance, and the second term, which goes to zero as the mean predictions of all models become identical, signifies epistemic variance.
Consequently, as shown in Fig. \ref{fig:si_training}, the transmission power $p^*_n(\bH)$ is predicted as $\hat\mu_n(\bH)$, and its corresponding uncertainty is estimated as $\hat\sigma_n^2(\bH)$ by employing the DNN models.

\subsection{Self-Improving Deep Learning for Interference Management}

The quantified uncertainty of predictions can be used to improve deep learning \cite{mukherjee2020uncertainty,Wang_Wang_Huang_Gao_Zhou_2023}.
Commonly, typical uncertainty-aware deep learning methods in machine learning literature focus on maximizing prediction accuracy by leveraging uncertainty.
However, in the realm of wireless communications, the goal of deep learning-based interference management is not to precisely estimate the power control solution, but to improve the system performance, specifically by maximizing the sum-rate as in \eqref{eqn:problem}.
Hence, when designing self-improving deep learning, it is insufficient to simply consider the quantified uncertainty of the predicted solution as in \eqref{eqn:quantified_uncertainty} for effectively pursuing the goal of interference management.

To address this issue, we propose a novel framework for self-improving interference management that elaborately considers actual system performance in light of the quantified uncertainty in deep learning.
In the framework, first during the training stage, the interference management and corresponding uncertainty quantification models are trained as described in Section \ref{sec:UQ_ICIC}.
In the subsequent self-improving stage, for each instance of $\bH$, the framework obtains the predicted solution $\hat{\bp}(\bH)$ and its quantified uncertainty.
Then, based on the quantified uncertainty, the credibility of the prediction is qualified.
If the predicted solution is deemed credible, it is used as is.
If not, the predicted solution is refined using existing optimization-based methods,\footnote{Note that the predicted solution itself is a plausible solution. Hence, enhancing this predicted solution is substantially more cost-efficient compared to directly seeking the optimal solution from scratch. This efficiency is evident in our empirical experiments. In one sample case, using the predicted solution $\hat{\bp}^*$ as an initial point significantly reduces computational time. The WMMSE algorithm requires only $0.096$\,ms of CPU time when starting with $\hat{\bp}^*$, as opposed to $8.628$\,ms when no initial point is provided.} and this improved solution is then used to further enhance the DNN models, as illustrated in Fig. \ref{fig:si_self_improving}.

To qualify the credibility, we need to design a qualifying criterion that can assess how much the predicted solution is close to the optimal one in terms of the system performance.
For design, we first construct a confidence interval for each predicted transmission power, using the standard deviation of epistemic uncertainty $\hat{\sigma}_{n,{\rm epi}}$, as
\begin{equation}
\label{eqn:confidence_interval}
\tilde{\mathcal{P}}_n=[\hat{p}_n^L,\hat{p}_n^U]=[\hat{p}^*_n-\alpha\hat{\sigma}_{n,{\rm epi}}, \hat{p}^*_n+\alpha\hat{\sigma}_{n,{\rm epi}}],
\end{equation}
where $\alpha$ is the control parameter that determines the confidence level.
With the confidence interval, we can define an uncertainty-aware feasible set of solutions as $\tilde{\mathcal{P}}=\prod_{n=1}^N\tilde{\mathcal{P}}_n$.
Since the optimal solution is likely to belong to the uncertainty-aware feasible set, comparing the sum-rate with the predicted solution (i.e., $\hat{R}=R(\mathbf{H},\hat{\mathbf{p}})$) and the maximum sum-rate achievable within the set $\tilde{\mathcal{P}}$ (i.e., $\tilde{R}=\max_{\bp\in\tilde{\mathcal{P}}}R(\mathbf{H},\mathbf{p})$) is a straightforward way to design the criterion.
If $\hat{R}$ is sufficiently close to $\tilde{R}$, the predicted solution $\hat{\bp}^*$ is considered credible from the perspective of system performance, and if not, it requires enhancement.
However, identifying the maximum achievable sum-rate within the feasible set is equivalent to solving the power allocation problem for interference management.
Consequently, the comparison of $\hat{R}$ and $\tilde{R}$ is not practical.


Instead, we design a qualifying criterion to evaluate whether the uncertainty-aware feasible set is sufficiently small to minimally affect the sum-rate.
This approach is plausible for achieving a sum-rate close to the optimal one, as the feasible set likely contains the optimal solution; the sum-rate achieved by the predicted solution will be close to the optimal one if the feasible set is sufficiently small.
To this end, we define the maximal and minimal interference solutions as $\hat{\bp}^U=[\hat{p}_n^U]_{n\in\mathcal{N}}$ and $\hat{\bp}^L=[\hat{p}_n^L]_{n\in\mathcal{N}}$, respectively, and calculate their corresponding sum-rates as $\hat{R}^U=R(\mathbf{H},\hat{\mathbf{p}}^U)$ and $\hat{R}^L=R(\mathbf{H},\hat{\mathbf{p}}^L)$, respectively.
Then, we design a qualifying criterion involving $\hat{R}$, $\hat{R}^U$, and $\hat{R}^L$ as 
\begin{equation}
	\label{eqn:criterion}
	\frac{\textrm{maxdist}(\hat R^U,\hat R^L,\hat R)}{\hat R}\leq \epsilon,
\end{equation}
where $\textrm{maxdist}(a,b,c)=\max\{\lvert a-b\rvert, \lvert a-c\rvert, \lvert b-c\rvert\}$, and $\epsilon$ is a criterion value that controls the criterion's sensitivity.
This criterion evaluates whether the uncertainty-aware feasible set is small enough so that the sum-rates achieved by the extreme solutions within the feasible set and the predicted solution (i.e., $\hat{R}^U$, $\hat{R}^L$, and $\hat{R}$) appear similar.
If this criterion is satisfied, the predicted solution $\hat{\bp}^*$ is considered credible since $\tilde{R}$ is expected to be close to $\hat{R}$.
On the other hand, if not due to a large uncertainty-aware feasible set, the prediction solution requires enhancement.


\subsection{Description of Self-Improving Interference Management Framework}

\algnewcommand{\LineComment}[1]{\Statex \(\triangleright\) #1}
\setlength{\textfloatsep}{5pt}
\begin{algorithm}[t]
	\caption{Self-Improving Interference Management Framework}\label{alg:algorithm}
	\footnotesize
	\begin{algorithmic}[1]
	\LineComment{\textit{Training Stage} (Fig. \ref{fig:si_training})}
	\State Generate dataset for interference management $\mathcal{D}$ with the target algorithm
	\State Build $M$ DNN models that have outputs for the mean and variance of $p_n^*$'s
	\State Initialize $\btheta_1,\btheta_2,...,\btheta_M$ randomly
	\For{$m=1,2,...,M$}
	\State Training $\btheta_m$ by minimizing $\mathcal{L}_{n,\btheta_m}$ in \eqref{eqn:loss_function} w.r.t. $\btheta_m$
	\EndFor
	\vspace{0.7em}
	\LineComment{\textit{Self-Improving Stage} (Fig. \ref{fig:si_self_improving})}
	\While{TRUE}
	\State Interference management request with instance $\bH$ arrives
	\State Compute $\hat{\bp}^*(\bH)$ and $\hat{\sigma}_{n,{\rm epi}}(\bH)$'s using $\btheta_m$'s
	\State Obtain the uncertainty-aware feasible set $\tilde{\mathcal{P}}(\bH)$ with $\tilde{\mathcal{P}}_n$'s in \eqref{eqn:confidence_interval}
	\If{The qualifying criterion in \eqref{eqn:criterion} is TRUE}
	\State Transmit using $\hat{\bp}^*(\bH)$
	\Else
	\State Enhance $\hat{\bp}^*(\bH)$ to $\bp^*(\bH)$ and transmit using it
	\State Store the enhanced solution into the self-improving dataset $\mathcal{D}_{\rm SI}$
	\EndIf
	\If{$|\mathcal{D}_{\rm SI}|\geq N_{\rm SI}$}
	\State Improve $\btheta_m$'s using $\mathcal{D}_{\rm SI}$ and clear $\mathcal{D}_{\rm SI}$\label{alg:line:self_improving}
	\EndIf
	\EndWhile
	\end{algorithmic}
\end{algorithm}

Here, we describe the self-improving interference management framework, illustrated in Fig. \ref{fig:self_improving_deep_learning} and detailed in Algorithm \ref{alg:algorithm}.
First, in the training stage (Fig. \ref{fig:si_training}), a dataset for interference management is generated using the target algorithm and channel instances as $(\bH,\bp^*(\bH))$'s (line 1).
We denote this dataset by $\mathcal{D}$.
For prediction and uncertainty quantification, we construct $M$ DNN models, each outputting the means and variances of the transmission power $p^*_n$'s as $g_{\btheta_m}(\bH)=[\hat{\mu}_{n,\btheta_m}(\bH),\hat{\sigma}^2_{n,\btheta_m}(\bH)]_{n\in\mathcal N}$ (line 2).\footnote{It is worth noting that any kind of DNN model capable of producing mean and variance as outputs can be used for self-improving interference management.}
The weights of each DNN model $m$, $\btheta_m$, are randomly initialized (line 3) and trained to minimize the negative log-likelihood loss function in \eqref{eqn:loss_function} with respect to $\btheta_m$ (lines 4--6).
During this training stage, various learning techniques such as cross-validation or early stopping can be employed.

In the self-improving stage (Fig. \ref{fig:si_self_improving}), for each arrival of interference management request with instance $\bH$, the predicted solution $\hat{\bp}^*(\bH)$ and its epistemic standard deviation $\hat{\sigma}_{n,{\rm epi}}(\bH)$ are computed using deep ensemble with $M$ DNN models as in Section \ref{sec:UQ_ICIC} (lines 8--9).
Using these calculations, the uncertainty-aware feasible set $\tilde{\mathcal{P}}(\bH)$ is constructed with $\tilde{\mathcal{P}}_n$'s in \eqref{eqn:confidence_interval} (line 10).
Then, the qualifying criterion in \eqref{eqn:criterion} is examined with $\tilde{\mathcal{P}}$; if the criterion is met, the predicted solution $\hat{\bp}^*(\bH)$ is directly used for transmission (line 12). If not, it is enhanced by using the target algorithm starting from the initial point $\hat{\bp}^*(\bH)$, and the enhanced solution is then used for transmission (line 14).
The enhanced solution (i.e., $\hat{\bp}^*$) is collected into a self-improving dataset $\mathcal{D}_{\rm SI}$ (line 15).
The DNN models, $\btheta_m$'s, are improved using $\mathcal{D}_{\rm SI}$ if the number of collected enhanced solutions is large enough to improve the models, i.e., $\lvert\mathcal{D}_{\rm SI}\rvert \ge N_{\rm SI}$, where $\lvert\mathcal{X}\rvert$ denotes the cardinality of set $\mathcal{X}$, and $N_{\rm SI}$ denotes the required number of samples for model improvement.
The self-improving dataset is cleared once it is used (lines 17--19).
The model improvement can be achieved either by simply retraining the DNN models with the combined dataset $\mathcal{D}\cup\mathcal{D}_{\rm SI}$ or by adopting continual learning techniques \cite{wang2023comprehensive}.

\section{Experimental Results}
\label{sec:simulation_results}
In this section, we provide experimental results to demonstrate the effectiveness of our self-improving interference management framework.
To this end, we implement the self-improving interference management framework (SI-DNN) in Python using a library for uncertainty quantification in deep learning called Fortuna \cite{detommaso2023fortuna}.
We set $\alpha$ to $1.96$ (for a $95$\% confidence interval and $\epsilon$ to $0.2$.
We also implement a DNN-based interference management without self-improving (DNN).
For those approaches, we consider a DNN architecture consisting of $5$ hidden layers with $1000$, $1000$, $500$, $500$, $100$ units, respectively. Each DNN model is trained by using the Adam optimizer with learning rate $10^{-3}$ and batchsize 100.
As a target algorithm of deep learning models, we consider the WMMSE algorithm as described in \cite{sun2018learning}. Hence, the sum-rate performance of WMMSE serves as an upper bound.
Furthermore, for the purpose of comparison, we also consider random power allocation (RandPower) and maximum power allocation (MaxPower).
Additionally, we adopt a network setup with $10$ transmission links considered in \cite{sun2018learning}. Each channel coefficient is generated using a pathloss coefficient of $-3.76$ and Rayleigh fading.

To clearly demonstrate the self-improvement of the models, we consider a scenario involving two topologies, Topology A and Topology B, each with different locations of transmitters and receivers.
In each topology, the distances between the transmitters and receivers within the same transmission links are randomly chosen from the interval $[10,15]$\,m, while the distances between the transmitters and receivers of different transmission links are randomly chosen from the interval $[10,20]$\,m. During the training stage, the DNN models are trained using a training dataset generated exclusively from Topology A. However, during the testing (self-improving) stage, the trained models are tested using a testing dataset generated from both Topologies A and B.
%
Consequently, it is more likely to encounter less reliable predictions (i.e., untrustworthy solutions) in Topology B than in Topology A.
Through this scenario, we can clearly show how the self-improving interference management addresses the issue of untrustworthy samples.

\begin{table}[!t]
\begin{center}
\caption{Comparison of sum-rate performance.}
\label{table:result}
\begin{tabular}{c|ccc}
\hline
\bfseries Alg. & \bfseries Total & \bfseries Topology A & \bfseries Topology B \\ \hline\hline
WMMSE       & 3.925 (100.00\%)  &    3.831 (100.00\%)    &    4.018 (100.00\%)    \\
SI-DNN      &  3.845 (97.98\%)  &    3.767 (98.33\%)     &    3.924 (97.65\%)     \\
DNN         &  3.483 (88.74\%)  &    3.599 (93.96\%)     &    3.366 (83.77\%)     \\
MaxPower    &  2.440 (62.17\%)  &    2.276 (59.41\%)     &    2.604 (64.81\%)     \\
RandPower   &  2.072 (52.80\%)  &    1.971 (51.44\%)     &    2.174 (54.10\%)     \\ \hline
\end{tabular}
\end{center}
\end{table}

We first provide the sum-rate performance of the algorithms in Table \ref{table:result}.
From the table, we can observe that the sum-rate achieved by DNN for Topology A closely approximates that of WMMSE, but this similarity is not as pronounced for Topology B.
This discrepancy arises because certain samples from Topology B are difficult to be effectively addressed by exploiting only the knowledge acquired from Topology A.
In contrast, SI-DNN consistently achieves the sum-rates that are closely aligned with WMMSE, irrespective of the underlying topologies.
This clearly demonstrates its ability to identify untrustworthy predictions and improve upon them.

\begin{figure*}[!t]
\centering
\subfloat[Total instances.]{\includegraphics[width=.25\linewidth]{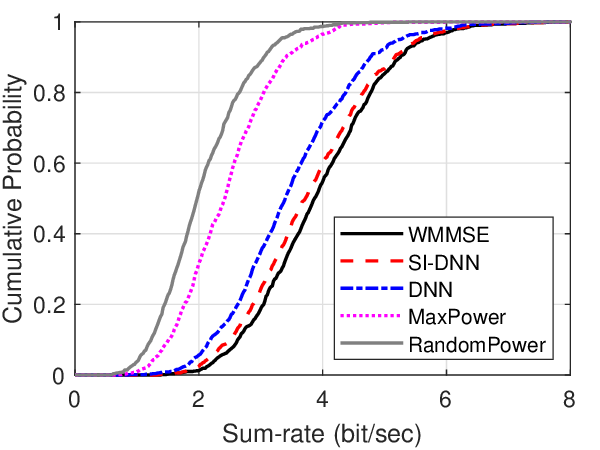}%
\label{fig:cdf_tot}}
\hfil
\subfloat[Instances from Topology A.]{\includegraphics[width=.25\linewidth]{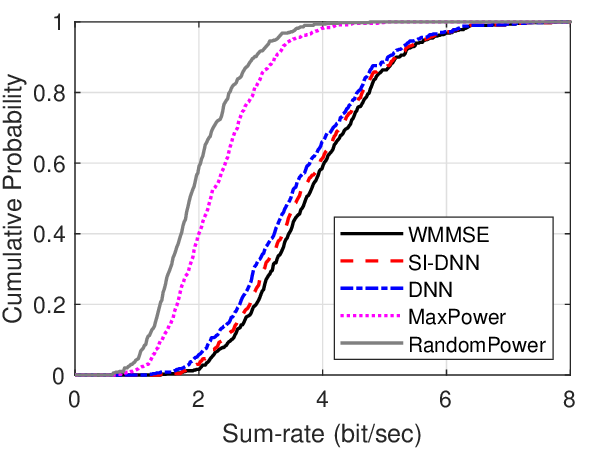}%
\label{fig:cdf_t1}}
\hfil
\subfloat[Instances from Topology B.]{\includegraphics[width=.25\linewidth]{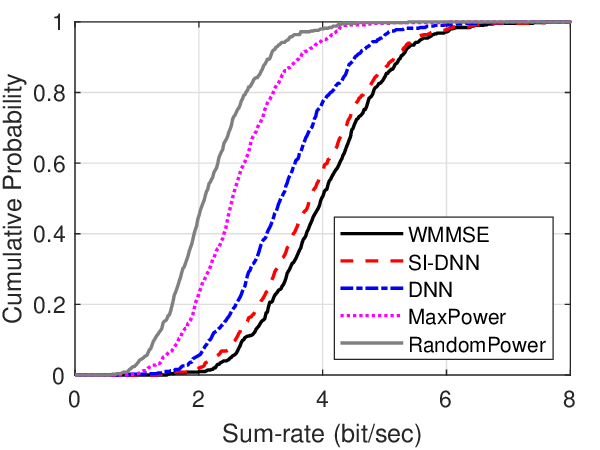}%
\label{fig:cdf_t2}}
\caption{The CDFs of the sum-rates achieved by different algorithms.}
\label{fig:cdf}
\end{figure*}

For further investigation, we provide the cumulative distribution function (CDF) of the sum-rates achieved by each algorithm in Fig. \ref{fig:cdf}.
From the figure, we can observe that the CDF curve of SI-DNN closely aligns with that of WMMSE, regardless of the underlying topologies.
As indicated in Table \ref{table:result}, the CDF curve of DNN closely aligns with that of WMMSE only for instances from Topology A.
Notably, we can see that SI-DNN exhibits cumulative probabilities lower than DNN across the entire sum-rate range, which implies that SI-DNN consistently outperforms DNN with a uniform gain.

\begin{figure}[!t]
\centering
\subfloat[Instance from Topology A.]{\includegraphics[width=.485\linewidth]{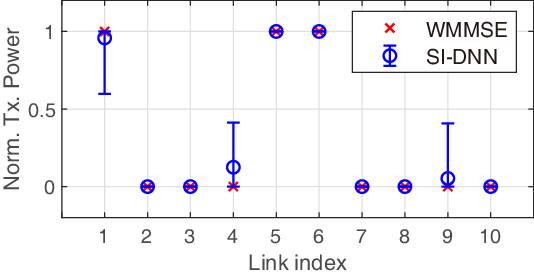}%
\label{fig:ctrue_uq}}
\hfil
\subfloat[Instance from Topology B.]{\includegraphics[width=.485\linewidth]{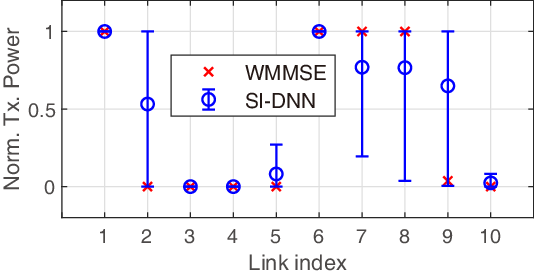}%
\label{fig:cfalse_uq}}
\caption{The predicted solutions and confidence intervals computed by SI-DNN for instances from different topologies.}
\label{fig:uncertainty_quantification}
\end{figure}

In Fig. \ref{fig:uncertainty_quantification}, we provide the predicted solutions and corresponding confidence intervals computed by SI-DNN for an instance from each topology to comprehend the uncertainty quantification.
The figures describe the normalized transmission power of each link within the rage of $[0,1]$.
In Fig. \ref{fig:ctrue_uq}, the predicted solution for the instance from Topology A is provided, and it is assessed as trustworthy.
We can observe that the confidence interval is narrow overall, and the solution is predicted well.
On the other hand, in Fig. \ref{fig:cfalse_uq}, the predicted solution for the instance from Topology B is provided, and it is assessed as untrustworthy.
In this instance, we note that the confidence interval is wide for many links, and indeed, the predictions for those links are not precise.

\begin{figure}[!t]
\centering
\subfloat[Enhancing rate.]{\includegraphics[width=.485\linewidth]{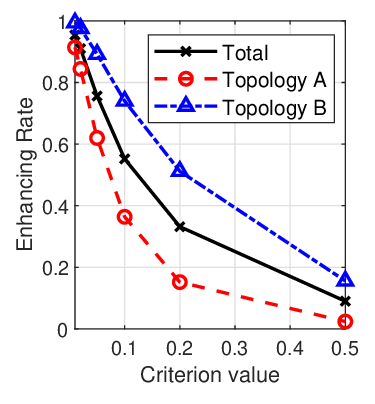}%
\label{fig:c_var_si}}
\hfil
\subfloat[Sum-rate.]{\includegraphics[width=.485\linewidth]{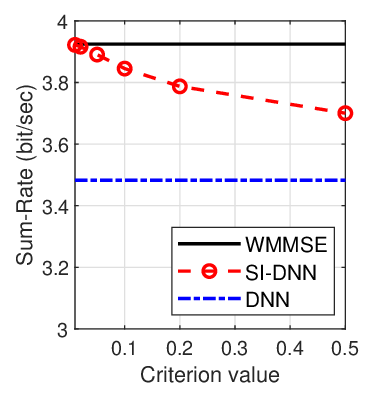}%
\label{fig:c_var_rate}}
\caption{The performance of the algorithms varying the criterion value $\epsilon$.}
\label{fig:c_var}
\end{figure}

To show the impact of the criterion value $\epsilon$ in \eqref{eqn:criterion}, we provide the enhancing rate and sum-rate of the algorithms while varying $\epsilon$ in the set $\{0.01,0.02,0.1,0.2,0.5\}$ in Fig. \ref{fig:c_var}.
The enhancing rate quantifies the proportion of predictions that undergo enhancement.
As the criterion value increases, the qualifying criterion becomes more easily met, and thus, enhancing predictions for self-improving occurs more rarely, as shown in Fig. \ref{fig:c_var_si}.
Accordingly, in Fig. \ref{fig:c_var_rate}, the sum-rate of SI-DNN decreases as the criterion value increases.
%
Furthermore, from Fig. \ref{fig:c_var_si}, we can observe that the predictions for Topology B are notably more frequently enhanced compared to those for Topology A.
This clearly demonstrates that SI-DNN effectively assesses the trustworthy of the predictions, given that the DNN model is trained exclusively on the dataset from Topology A.

\begin{figure}[!t]
\centering
\subfloat[Enhancing rate.]{\includegraphics[width=.485\linewidth]{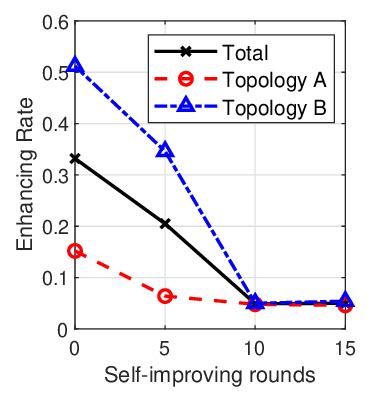}%
\label{fig:rounds_si}}
\hfil
\subfloat[Sum-rate.]{\includegraphics[width=.485\linewidth]{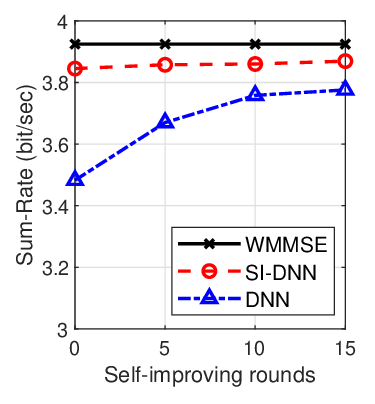}%
\label{fig:rounds_rate}}
\caption{The performance of the algorithms as self-improving progresses.}
\label{fig:rounds}
\end{figure}

To show the effectiveness of self-improving, we provide the enhancing rate and sum-rate of the algorithms according to the self-improving rounds in Fig. \ref{fig:rounds}, where each self-improvement round corresponds to executing line \ref{alg:line:self_improving} in Algorithm \ref{alg:algorithm} once.
For self-improving, we set $N_{\rm SI}$ to $1000$ to conduct one round of self-improving and assume that DNN is re-trained using the dataset $\mathcal{D}\cup\mathcal{D}_{\rm SI}$ for each round.
As self-improving progresses, the DNN model is improved by using the enhanced samples for which it previously predicted untrustworthy solutions.
Hence, the enhancing rate decreases, as shown in Fig. \ref{fig:rounds_si}.
From the figure, we can observe that the enhancing rate of SI-DNN converges after $10$ rounds.
We can explain that this convergence in the enhancing rate can be attributed to limitations in the DNN model, such as its representational capacity and generalization capability.
From Fig. \ref{fig:rounds_rate}, we can see that the sum-rate of DNN experiences effective improvement with the enhanced samples.
This clearly demonstrates SI-DNN's ability to proficiently identify untrustworthy predictions.

\section{Conclusion}
In this paper, we have proposed the self-improving interference management framework.
If the solution provided by the deep learning-based algorithm is deemed untrustworthy, it enhances the solution using optimization-based algorithms in a complementary manner, and these enhanced solutions are utilized to refine the DNN model.
To this end, we have proposed the qualifying criterion for solutions based on quantified uncertainties while considering system performance.
Through the simulation results, we have demonstrated that the proposed framework adeptly identifies untrustworthy solutions and efficiently enhances the DNN model for interference management.

\bibliographystyle{IEEEtran}
\bibliography{IEEEabrv,main_manuscript.bib}

\end{document}